\documentclass{article}

\usepackage[final,nonatbib]{arxiv}

\usepackage{todonotes}
\usepackage[utf8]{inputenc} % allow utf-8 input
\usepackage[T1]{fontenc}    % use 8-bit T1 fonts
\usepackage{hyperref}       % hyperlinks
\usepackage{url}            % simple URL typesetting
\usepackage{booktabs}       % professional-quality tables
\usepackage{amsfonts}       % blackboard math symbols
\usepackage{nicefrac}       % compact symbols for 1/2, etc.
\usepackage{microtype}      % microtypography
\usepackage{tabularx, booktabs,adjustbox}
\usepackage{multirow}
\usepackage{caption}
\usepackage{subcaption}
\usepackage{array}
\usepackage{authblk}
\usepackage{float}
\usepackage{multirow}
\usepackage{footnote}
\usepackage[numbers]{natbib}

\begin{document}

\title{Machine Friendly Machine Learning: \\ Interpretation of Computed Tomography Without Image Reconstruction}

\author[1,2]{Hyunkwang Lee}
\author[2]{Chao Huang}
\author[2]{Sehyo Yune}
\author[2]{Shahein H. Tajmir}
\author[2]{Myeongchan Kim}
\author[2]{Synho Do}

\affil[1]{School of Engineering and Applied Sciences, Harvard University, Cambridge, MA}
\affil[2]{Department of Radiology, Massachusetts General Hospital, Boston, MA}

\maketitle

\begin{abstract}
Recent advancements in deep learning for automated image processing and classification have accelerated many new applications for medical image analysis. However, most deep learning applications have been developed using reconstructed, human-interpretable medical images. While image reconstruction from raw sensor data is required for the creation of medical images, the reconstruction process only uses a partial representation of all the data acquired. Here we report the development of a system to directly process raw computed tomography (CT) data in sinogram-space, bypassing the intermediary step of image reconstruction. Two classification tasks were evaluated for their feasibility for sinogram-space machine learning: body region identification and intracranial hemorrhage (ICH) detection. Our proposed SinoNet performed favorably compared to conventional reconstructed image-space-based systems for both tasks, regardless of scanning geometries in terms of projections or detectors. Further, SinoNet performed significantly better when using sparsely sampled sinograms than conventional networks operating in image-space. As a result, sinogram-space algorithms could be used in field settings for binary diagnosis testing, triage, and in clinical settings where low radiation dose is desired. These findings also demonstrate another strength of deep learning where it can analyze and interpret sinograms that are virtually impossible for human experts.
\end{abstract}

\section{Introduction}

Continued rapid advancements in algorithms and computer hardware have accelerated progress in automated computer vision and natural language processing. By combining these two factors with the availability of well-annotated large datasets, significant advances have emerged from automated medical image interpretation for the detection of disease and critical findings \cite{esteva2017dermatologist, gulshan2016development, chilamkurthy2018deep}. The application of deep learning has the potential to increase diagnostic accuracy and reduce delays in diagnosis and treatment for better patient outcomes \cite{thrall2018artificial}. Deep learning techniques are not limited to image analysis, but they also can improve image reconstruction for magnetic resonance imaging (MRI) \cite{wang2016accelerating, zhu2018image}, computed tomography (CT) \cite{xie2018artifact, jin2017deep}, and photoacoustic tomography (PAT) \cite{antholzer2018deep}. Deep learning now is a feasible alternative to well-established analytic and iterative methods of image reconstruction \cite{wang2018image, do2014high, do2014sinogram, do2013iterative, do2011decomposition}.

However, most prior work using deep learning algorithms has focused on image analysis of reconstructed images or as an alternative approach to image reconstruction. Despite this human centric approach, there is no reason that deep learning algorithms must function in image-space. Since all the information in the reconstructed images is present in the raw measurement data, deep learning models could potentially derive features directly from raw data in sinogram-space without intermediary image reconstruction, with possibly even better performance than models trained in image-space. In this study, we determined the feasibility analyzing computed tomography (CT) projection data - sinograms - through a deep learning approach for human anatomy identification and pathology detection. We proposed a customized convolutional neural network (CNN) called SinoNet, optimized it for interpreting sinograms, and demonstrated its potential by comparing its performance to pre-existing system based on other CNN architectures using reconstructed CT images. This approach accelerates edge computing by making it possible to identify critical findings rapidly from the raw data without time-consuming image reconstruction processes. In addition, this could enable us to develop simplified scanner hardware for the direct detection of critical findings through SinoNet alone.

\section{Results}
\subsection{Experimental design}

We retrieved 200 contiguous whole body CT datasets from combined positron emission tomography-computed tomography (PET/CT) examinations for body part recognition and 720 non-contrast head CT scans for intracranial hemorrhage (ICH) detection with IRB approval from the picture archiving and communication systems at our quaternary referral hospital. Axial slices in the 200 whole body scans were annotated as sixteen different body regions by a physician, and slices of the 720 head scans were annotated with the presence of hemorrhage by a panel of five neuroradiologists by consensus (Methods). We evaluated twelve different classification models developed by training Inception-v3 \cite{szegedy2016rethinking} on reconstructed CT images and SinoNet with sinograms (Table \ref{table:models}, Methods). The reconstructed CT images containing Hounsfield units (HU) were converted to scaled linear attenuation coefficients (LAC). Two-dimensional (2D) parallel-beam Radon transform was applied to the LAC slices (512x512 pixels) to generate a fully-sampled sinogram with 360 projections and 729 detector pixels (\textit{sino360x729}), which was then uniformly subsampled in the horizontal direction (projection views) and averaged in vertical direction (detector pixels) by factors of 3 and 9 to obtain moderately sampled sinograms with 120 views by 240 pixels (\textit{sino120x240}) and sparsely sampled sinograms with 40 views by 80 pixels (\textit{sino40x80}). 

Original CT images were used as fully sampled reconstructed images (\textit{recon360x729}), and images reconstructed from the sparse sinograms (\textit{recon120x240} and \textit{recon40x80}) were generated using a deep learning approach (FBPConvNet \cite{jin2017deep}) followed by a conversion from LAC to HU. Reconstructed CT images and sinograms with predefined window-level settings were created to evaluate the effect of windowing: \textit{wrecon360x729}, \textit{wrecon120x240}, \textit{wrecon40x80}; and \textit{wsino360x729}, \textit{wsino120x240}, \textit{wsino40x80} (Methods). Based on the scanning geometries and window-level settings described above, 12 CNN models were evaluated: 6 were developed by training Inception-v3 \cite{szegedy2016rethinking} with reconstructed CT images and the other 6 were obtained by training SinoNet with sinograms (Table \ref{table:models}, Methods). Data for body part recognition was randomly split into training, validation, and test sets with balanced genders: 140 scans in training, 30 in validation, and 30 in testing. A similar dataset breakdown was performed for ICH detection with 478 scans in training, 121 in validation, and 121 in testing. Details of data preparation, CNN architecture, sinogram generation, and image reconstruction are described in Methods.

\begin{table}[!htbp]
\centering
\scalebox{0.8}{
\begin{tabular}{l|l|l}
    \toprule
    % \hline
    \multicolumn{1}{c|}{\textbf{Fully sampled}} & 
    \multicolumn{1}{c|}{\textbf{Moderately sampled}} & 
    \multicolumn{1}{c}{\textbf{Sparsely sampled}} \\
    
    \multicolumn{1}{c|}{360 projections and 729 detectors} & 
    \multicolumn{1}{c|}{120 projections and 240 detectors} & 
    \multicolumn{1}{c}{40 projections and 80 detectors} \\
    % \hline
    \midrule
    
    \textbf{I1}: \textit{recon360x729} (original CT) &
    \textbf{I3}: \textit{recon120x240} &
    \textbf{I5}: \textit{recon40x80} \\
    % \hline
    
    \textbf{S1}: \textit{sino360x729}&
    \textbf{S3}: \textit{sino120x240} &
    \textbf{S5}: \textit{sino40x80} \\
    % \hline
    
    \textbf{I2}: \textit{wrecon360x729} (windowed original CT) &
    \textbf{I4}: \textit{wrecon120x240} &
    \textbf{I6}: \textit{wrecon40x80} \\
    % \hline
   
    \textbf{S2}: \textit{wsino360x729}&
    \textbf{S4}: \textit{wsino120x240} &
    \textbf{S6}: \textit{wsino40x80} \\
    
    \bottomrule
\end{tabular}}
\caption{Summary of the 12 different models evaluated in this study.}
\label{table:models}
\end{table}

\begin{figure}[!htbp]
\begin{center}
        \includegraphics[scale=1.0]{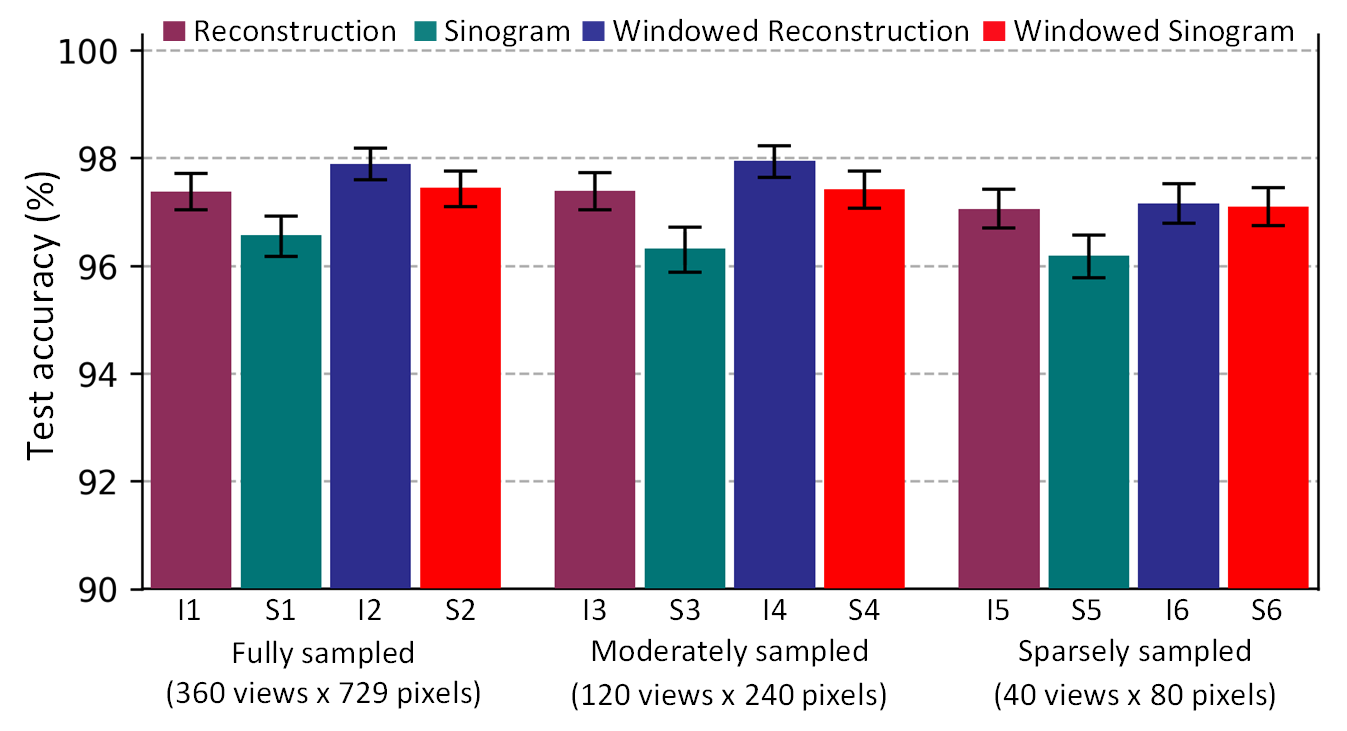}
        \caption{Performance of 12 different models trained on reconstruction images and sinograms with varying numbers of projections and detectors for body part recognition. 95\% confidence intervals (CIs) are indicated in black error bars. The purple and blue bars (I1-I6) compare the test accuracy of Inception-v3 trained with full dynamic range reconstructed images with abdominal window setting reconstructed images (window-level=40HU, window-width=400HU),. The green and red bars (S1-S6) compare the performance of SinoNet models trained with sinograms generated from full-range and windowed reconstructed images, respectively.}
        \label{fig:bodypartresults}
\end{center}
\end{figure}

\subsection{Results of body part recognition}

Figure \ref{fig:bodypartresults} shows test performance of the twelve different models for body part recognition. Models trained on fully sampled images had accuracies of 97.4\% in image-space, 96.6\% in sinogram-space, 97.9\% in windowed-image-space, and 97.4\% in windowed-sinogram-space. Moderately sampled images had model accuracies of 97.4\% in image-space, 96.3\% in sinogram-space, 97.9\% in windowed-image-space, and 97.4\% in windowed-sinogram-space. Sparsely sampled images had model accuracies of 97.1\% in image-space, 96.2\% in sinogram-space, 97.2\% in windowed-image-space, and 97.1\% in windowed-sinogram-space. These results imply that models trained and operating in image-space performed slightly better than sinogram-space (SinoNet) models for body part recognition, regardless of scanning geometry. Additionally, windowed input images consistently outperformed the ones with full-range images/sinograms.

\begin{figure}[!ht]
\begin{center}
\includegraphics[scale=1.0]{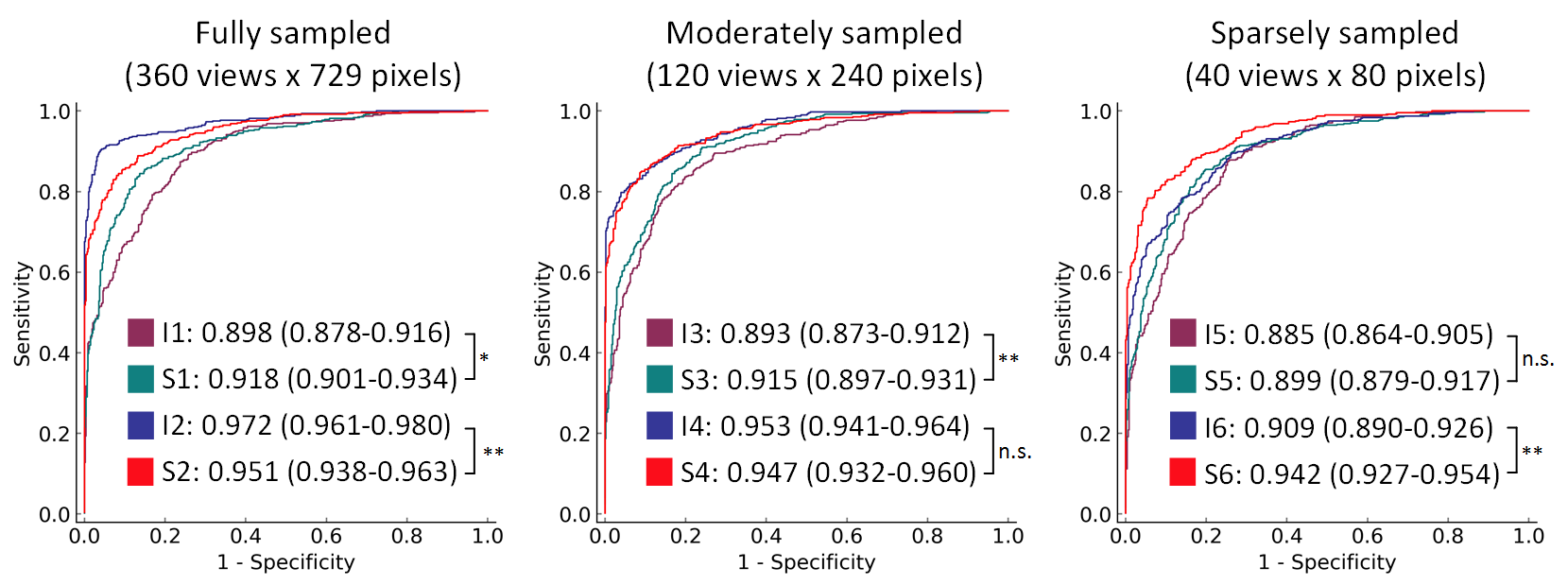}
\caption{ROC curves for performance of 12 different models trained with reconstruction images and sinograms with various sparsity configurations in numbers of projections and detectors. The purple and blue curves (I1-I6) correspond to performance of Inception-v3 trained with reconstruction images with a full dynamic range of HU values and brain window setting (window-level=50HU, window-width=100HU), respectively. The green and red curves (S1-S6) show performance of SinoNet models trained with sinograms generated from full-range and windowed reconstruction images, respectively. The areas under the curve (AUCs) for the 12 models are present in legends with their 95\% CIs. Statistical significance of the difference between AUCs of paired models (Ix - Sx) was evaluated. n.s., p>0.05; * p<0.05; ** p<0.01.}
\label{fig:ichresults}
\end{center}
\end{figure}

\subsection{Results of intracranial hemorrhage detection} 

Figure \ref{fig:ichresults} depicts receiver operating characteristic (ROC) curves, and the corresponding areas under the ROC curves (AUC) for the twelve different models of ICH detection. Models trained on fully sampled images had AUCs of 0.898 in image-space, 0.918 in sinogram-space, 0.972 in windowed-image-space, and 0.951 in windowed-sinogram-space. Moderately sampled images had model accuracies of 0.893 in image-space, 0.915 in sinogram-space, 0.953 in windowed-image-space, and 0.947 in windowed-sinogram-space. Sparsely sampled images had model accuracies of 0.885 in image-space, 0.899 in sinogram-space, 0.909 in windowed-image-space, and 0.942 in windowed-sinogram-space

\subsection{Comparison of SinoNet and Inception-v3 for analyzing sinograms}

Table \ref{table:sinoresults} details performance comparisons of Inception-v3 and SinoNet for interpreting fully-sampled sinograms (360 projection views and 729 detector pixels) for both body part recognition and ICH detection. SinoNet models significantly outperformed Inception-v3 models in both tasks. 

\begin{table}[!ht]
\centering
\scalebox{0.9}{
\begin{tabular}{c|c|c|c|c}
    \toprule
    % \hline
     & \multicolumn{2}{c|}{\textbf{Body part recognition (Accuracy)}} & 
    \multicolumn{2}{c}{\textbf{ICH detection (AUC)}} \\
    
    \textbf{Input} & \textbf{Inception-v3} & \textbf{SinoNet} & 
    \textbf{Inception-v3} & \textbf{SinoNet} \\
    % \hline
    \midrule
    
    \textbf{\textit{sino360x729}} &
    93.9\% (93.4\%-94.4\%) &
    96.6\% (96.2\%-96.9\%) &
    0.873 (0.849-0.895) &
    0.918\textsuperscript{*} (0.899-0.935) \\
    
    \textbf{\textit{sino120x240}} &
    93.5\% (93.0\%-94.0\%) &
    96.3\% (95.9\%-96.7\%) &
    0.874 (0.851-0.896) &
    0.915\textsuperscript{*} (0.897-0.932) \\
    
    \textbf{\textit{sino40x80}} &
    93.4\% (92.9\%-93.9\%) &
    96.2\% (95.8\%-96.6\%) &
    0.852 (0.828-0.876) &
    0.899\textsuperscript{*} (0.879-0.917) \\
    
    \bottomrule
\end{tabular}}
\caption{Comparison of Inception-v3 and SinoNet network performance when both networks are trained on full-range sinograms are varying sampling densities for body part recognition and intracranial hemorrhage (ICH) detection. Body part recognition is reported in accuracy. ICH detection as AUC. 95\% CIs in parentheses. * p<0.0001.}
\label{table:sinoresults}
\end{table}

\begin{figure}[!ht]
\begin{center}
        \includegraphics[width=\linewidth]{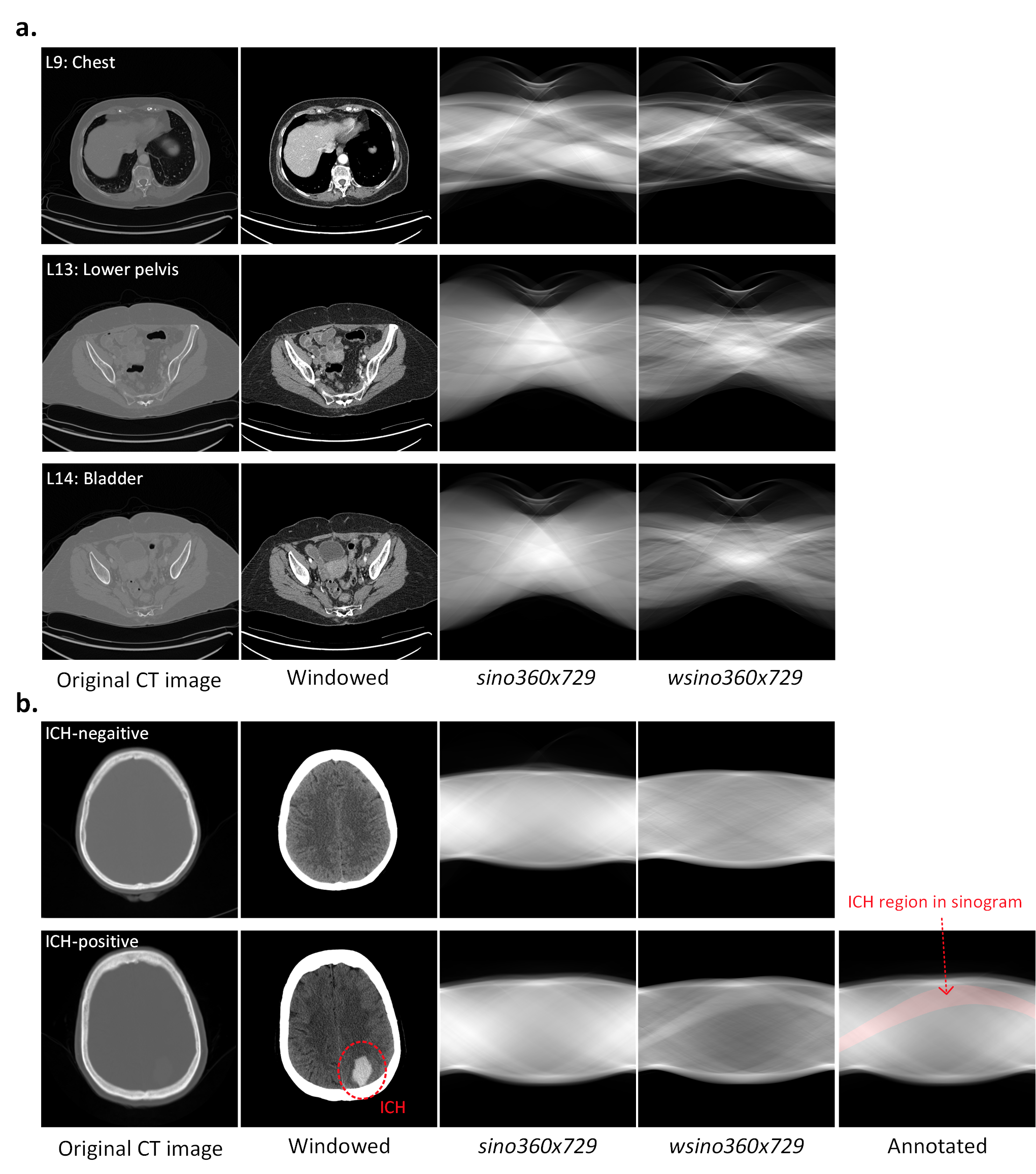}
        \caption{Examples of reconstructed images and sinograms with different labels for \textbf{a} body part recognition and \textbf{b} ICH detection. From left to right: original CT images, windowed CT images, sinograms with 360 projections by 729 detector pixels, and windowed sinograms 360x729. In the last row, an example CT with hemorrhage is annotated with a dotted circle in image-space with the region of interest converted into the sinogram domain using Radon transform. This area is highlighted in red on the sinogram in the fifth column.}
        \label{fig:examples}
\end{center}
\end{figure}

\section{Discussion}

We have demonstrated that models trained on sinograms can achieve similar performance when compared to models using conventional reconstructed images for body part recognition and ICH detection in all three scanning geometries, despite the fact that the measurement data are not interpretable to humans. SinoNet, when trained with sinograms, has comparable performance with that of Inception-v3 when trained with reconstructed CT images for body part recognition, regardless of the number of projection views or detectors. For ICH detection, SinoNet trained with full-range sinograms outperformed Inception-v3 trained with full dynamic range reconstructed images for all three scanning geometries, with SinoNet significantly outperforming Inception-v3 when using windowed, sparsely sampled images. By applying window settings similar to what a radiologist would use, network performance increased significantly due to the improved target to background (Figure \ref{fig:examples}) in both reconstructed images and in sinogram-space. As depicted in Figure \ref{fig:examples} (b), not only are the key features relevant to hemorrhage enhanced in the windowed CT image, but also in the windowed sinogram. 

SinoNet, a customized convolutional neural network, was developed for analyzing sinograms through customized Inception modules with multi-scale convolutional and pooling layers \cite{szegedy2016rethinking}. In SinoNet, the square convolutional filters in the original Inception module were replaced by various sized rectangular convolutional filters which include width-wise (projection dominant) and height-wise (detector dominant) filters. The customized architecture of SinoNet allowed for significantly improved performance in both body part recognition and ICH detection when compared with Inception-v3 models trained with sinograms, regardless of sampling density. These results imply that non-square filters may be effective in enabling models to learn the interplay between projection views and detector pixels from sinusoidal curves and to extract salient features from the sinogram domain for classification, a task thought to be impossible for human experts to grasp. This approach is similar to the one proposed for learning temporal and frequency features using rectangular convolution filters in spectrograms \cite{pons2016experimenting}. 

SinoNet, by operating in sinogram-space, can accelerate image interpretation for pathology detection as complex computations for image reconstruction are not required. SinoNet also excels when the projection data was moderately or sparsely sampled, maintaining its AUC at 0.942 on the hemorrhage detection task, while Inceptionv3 dropped from 0.972 to 0.909. Sparsely sampled datasets suggest that radiation dose could be markedly decreased with only a slight degradation in performance for sinogram-space algorithms. The number of projections linearly correlates with radiation dose, theoretically achieving 33\% and 89\% dose reductions for moderately and sparsely sampled data respectively. Similarly, by reducing the size and number of detectors required for diagnostic CT data, cheaper and simpler CT scanners can be created. At our institution, the average head CT has a CTDI\textsubscript{vol} of 50 mGy. Sparsely sampled data could have CTDI\textsubscript{vol} between 6 and 16 mGy. One possible use of this technique would be to use the sinogram model as a first-line screening tool in the field setting without image reconstruction, subsequently prioritizing a patient for potential stroke therapy given no evidence of intracranial hemorrhage. Subsequent full-dose CT could be used to confirm the interpretation from the sinogram method. Another possible use for this technique would be to create “smart-scanners” which allow the CT scanner to adjust the protocol and field of view based on the intended region of the body.

Although these results demonstrate the power of the sinogram based approach, several important areas of future investigation remain. Due to their unavailability, the sinograms used in this study were simulated by applying the 2D parallel-beam Radon transform to the reconstructed CT images rather than actual measurement data acquired from CT scanners. Improved simulation data could be acquired by accounting for other advanced projection geometries - cone-beam or fan-beam - and considering Poisson noise when generating projection data. Although SinoNet trained with windowed sinograms achieved comparable or better performance compared with windowed reconstructed images, windowed sinograms were generated from reconstructed images that were postprocessed with predefined window settings; generation of windowed sinograms directly from CT measurement data is not straightforward, but it could be implemented by using energy-resolving, photon-counting detectors from multi-energy CT imaging to acquire measurements in multiple energy bins \cite{mccollough2015dual}. Our work will need to be further validated by using raw data from clinical scanners as well as raw data from actual low-dose image acquisitions to see if performance remains robust despite increased image noise. 

\section{Methods}
\label{sec:methods}

This HIPAA-compliant retrospective study was conducted with the approval of our institutional review board and under a waiver of informed consent.

\subsection{Data collection and annotation}

\begin{figure}[!ht]
\begin{center}
        \includegraphics[width=\linewidth]{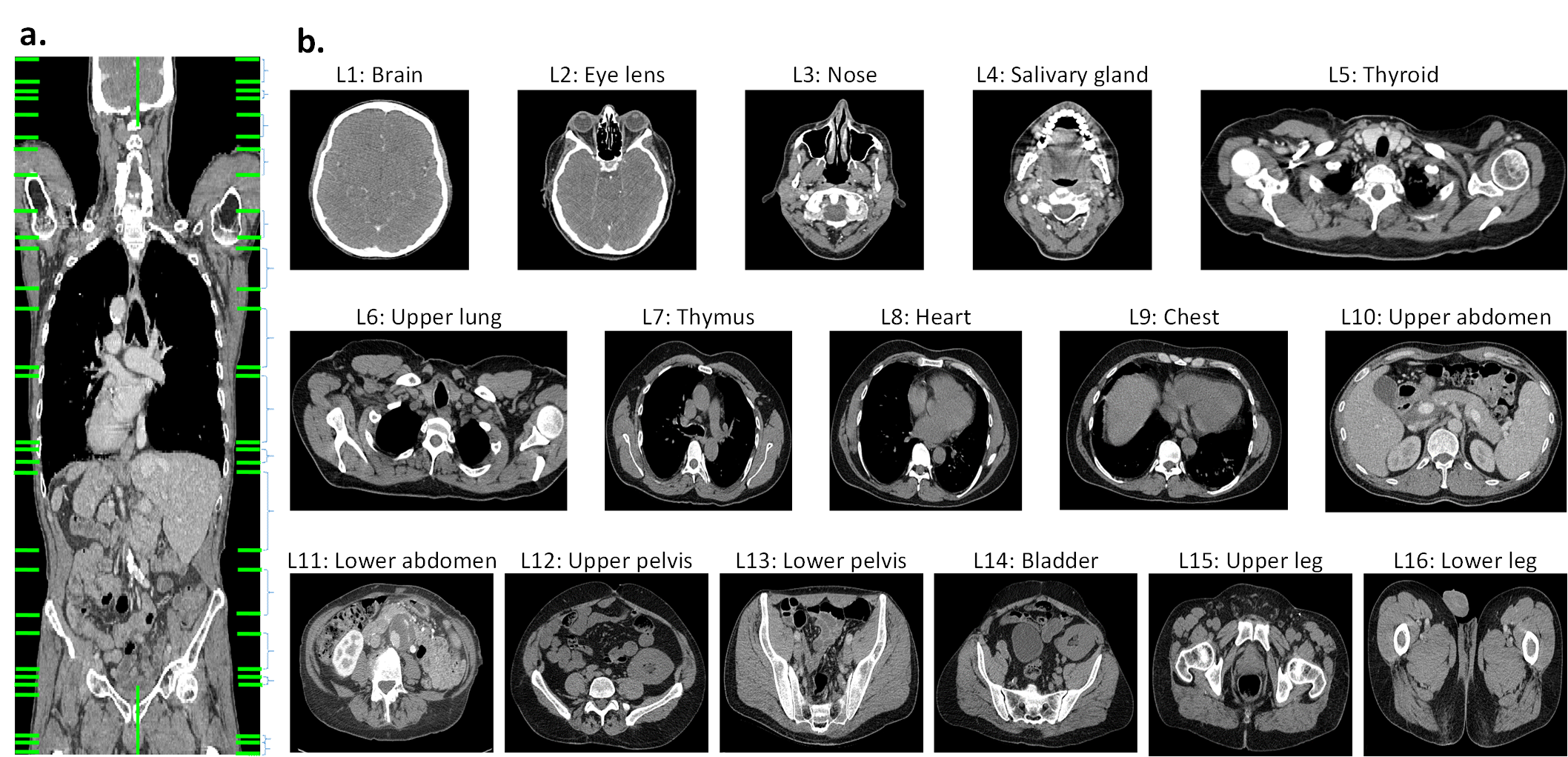}
        \caption{\textbf{a} A coronal view of a whole-body CT scan image with regions of each body part (annotated in green); \textbf{b} Representative CT images of 16 different body parts in axial view: L1=Brain, L2=Eye lens, L3=Nose, L4=Salivary gland, L5=Thyroid, L6=Upper lung, L7=Thymus, L8=Heart, L9=Chest, L10=Upper abdomen, L11=Lower abdomen, L12=Upper pelvis, L13=Lower pelvis, L14=Bladder, L15=Upper leg, L16=Lower leg.}
        \label{fig:bodypartlabel}
\end{center}
\end{figure}

\textbf{Body part recognition}: a total of 200 contrast-enhanced PET/CT examinations of head, neck, chest, abdomen, and pelvis for 100 female and 100 male patients were retrieved from our institutional Picture Archiving and Communication System (PACS) between May 2012 and July 2012. 56,334 axial slices in the CT scans were annotated as one of sixteen body regions by a physician (Figure \ref{fig:bodypartlabel}). 15\% of the total slices were randomly selected for use as validation data for hyperparameter tuning and model selection, 15\% as test data for performance evaluation, and the rest as training data for model development (Table \ref{table:bodypartdata}).

\textbf{Intracranial hemorrhage (ICH) detection}: a total of 720 5-mm non-contrast head CT scans were identified and retrieved from our PACS between June 2013 and July 2017. Every 5-mm thick axial slice (3,151 slices without ICH and 2,895 slices with ICH) was annotated by five board-certified neuroradiologists (blinded for review, 9 to 34 years experience) according to presence of ICH by consensus. The examinations included 201 cases without ICH and 519 cases with ICH, which were randomly split into train, validation, and test datasets at the case-level to ensure slices from the same case were not split across different datasets (Table \ref{table:ichdata}).

\begin{table}[!htbp]
\centering
\scalebox{0.9}{
\begin{tabular}{c|c|c|c}
    \toprule
    % \hline
     & \textbf{Train} & \textbf{Validation} & \textbf{Test} \\
    \midrule
    
    No. Cases & 140 (70F, 70M) & 30 (15F, 15M) & 30 (15F, 15M) \\
    \midrule

    No. Images & 39,472 & 8,383 & 8,479 \\
    % \hline
    \midrule
    
    \multicolumn{1}{l|}{L1: Head} & 1,980 & 483 & 435 \\
    \multicolumn{1}{l|}{L2: Eye lens} & 878 & 189 & 188 \\
    \multicolumn{1}{l|}{L3: Nose} & 1,449 & 309 & 323 \\
    \multicolumn{1}{l|}{L4: Salivary gland} & 1,803 & 361 & 349 \\
    \multicolumn{1}{l|}{L5: Thyroid} & 1,508 & 312 & 333 \\
    \multicolumn{1}{l|}{L6: Upper lung} & 1,632 & 345 & 392 \\
    \multicolumn{1}{l|}{L7: Thymus} & 3,213 & 727 & 672 \\
    \multicolumn{1}{l|}{L8: Heart} & 3,360 & 707 & 762 \\
    \multicolumn{1}{l|}{L9: Chest} & 4,647 & 914 & 935 \\
    \multicolumn{1}{l|}{L10: Upper abdomen} & 4,943 & 1,008 & 1,103 \\
    \multicolumn{1}{l|}{L11: Lower abdomen} & 1,736 & 342 & 368 \\
    \multicolumn{1}{l|}{L12: Upper pelvis} & 2,524 & 617 & 545 \\
    \multicolumn{1}{l|}{L13: Lower pelvis} & 2,230 & 563 & 422 \\
    \multicolumn{1}{l|}{L14: Bladder} & 3,144 & 609 & 766 \\
    \multicolumn{1}{l|}{L15: Upper leg} & 2,607 & 563 & 532 \\
    \multicolumn{1}{l|}{L16: Lower leg} & 1,818 & 334 & 354 \\
    
    \bottomrule
\end{tabular}}
\caption{Distribution of training, validation, and test datasets for body part recognition. F, Female; M, Male}
\label{table:bodypartdata}
\end{table}

\begin{table}[!htbp]
\centering
\scalebox{0.9}{
\begin{tabular}{c|c|c|c|c|c|c}
    \toprule
    % \hline
     & \multicolumn{2}{c|}{\textbf{Train}} & 
    \multicolumn{2}{c|}{\textbf{Validation}} &
    \multicolumn{2}{c}{\textbf{Test}} \\
    
     & \multicolumn{1}{c}{No. Cases} & \multicolumn{1}{c|}{No. Images} & 
    \multicolumn{1}{c}{No. Cases} & \multicolumn{1}{c|}{No. Images} &
    \multicolumn{1}{c}{No. Cases} & \multicolumn{1}{c}{No. Images} \\
    % \hline
    \midrule
    
    No ICH & 
    \multicolumn{1}{c}{141} & \multicolumn{1}{c|}{2,202} & 
    \multicolumn{1}{c}{30} & \multicolumn{1}{c|}{474} & 
    \multicolumn{1}{c}{30} & \multicolumn{1}{c}{475} \\ 

    ICH & 
    \multicolumn{1}{c}{337} & \multicolumn{1}{c|}{1,915} & 
    \multicolumn{1}{c}{91} & \multicolumn{1}{c|}{490} & 
    \multicolumn{1}{c}{91} & \multicolumn{1}{c}{475} \\
    \midrule

    Total & 
    \multicolumn{1}{c}{478} & \multicolumn{1}{c|}{4,117} & 
    \multicolumn{1}{c}{121} & \multicolumn{1}{c|}{964} & 
    \multicolumn{1}{c}{121} & \multicolumn{1}{c}{950} \\ 
    
    \bottomrule
\end{tabular}}
\caption{Distribution of training, validation, and test datasets for ICH detection.}
\label{table:ichdata}
\end{table}

\subsection{Sinogram generation}

Simulated sinograms were utilized in this study instead of raw data obtained by commercial CT scanners as this was a retrospective analysis and access to raw projection data from patient CT scans could not be retrieved. To generate simulated sinograms, the pixel values of 512x512 CT images stored in DICOM file were first converted into scaled linear attenuation coefficients (LACs). Any calculated negative LAC was leveled to zero under the assumption that it is physically impossible to have negative LACs, so this result must represent random noise. Subsequently, three different sinograms were generated based on the scaled LAC images. First, we computed sinograms with 360 projection views over 180 degrees and 729 detectors (\textit{sino360x729}), using the 2D parallel-beam Radon transform. sino360x729 were then used to produce sparser sinograms by uniformly subsampling projection views (in the horizontal direction) and averaging projection data from adjacent detectors (in the vertical direction) by factors of 3 and 9 to obtain sinograms with 120 projection views and 240 detectors (\textit{sino120x240}) and sinograms with 40 projection views and 80 detectors (\textit{sino40x80}), respectively (Figure \ref{fig:sinogramgen}). Sparser sinograms (\textit{sino40x80}, \textit{sino120x240}) were resized to 360x729 pixels using a bilinear interpolation to have a uniform resolution with the corresponding full-view sinograms (\textit{sino360x729}).

\begin{figure}[!htbp]
\begin{center}
        \caption{\textbf{a} Schematic of sinogram generation with 360 projection views and 729 detectors (\textit{sino360x729}) from original CT images (converted into linear attenuation coefficients). \textbf{b} Sparse sinograms were created from \textit{sino360x729} by downsampling in the horizontal dimension and signal averaging in the vertical dimension to simulate the effect of acquiring an image with 120 projection views and 240 detectors (\textit{sino120x240}) or an image with 40 projection views and 80 detectors (\textit{sino40x80})}
        \label{fig:sinogramgen}
\end{center}
\end{figure}

\subsection{Image reconstruction}

\begin{figure}[!ht]
\begin{center}
        \includegraphics[width=\linewidth]{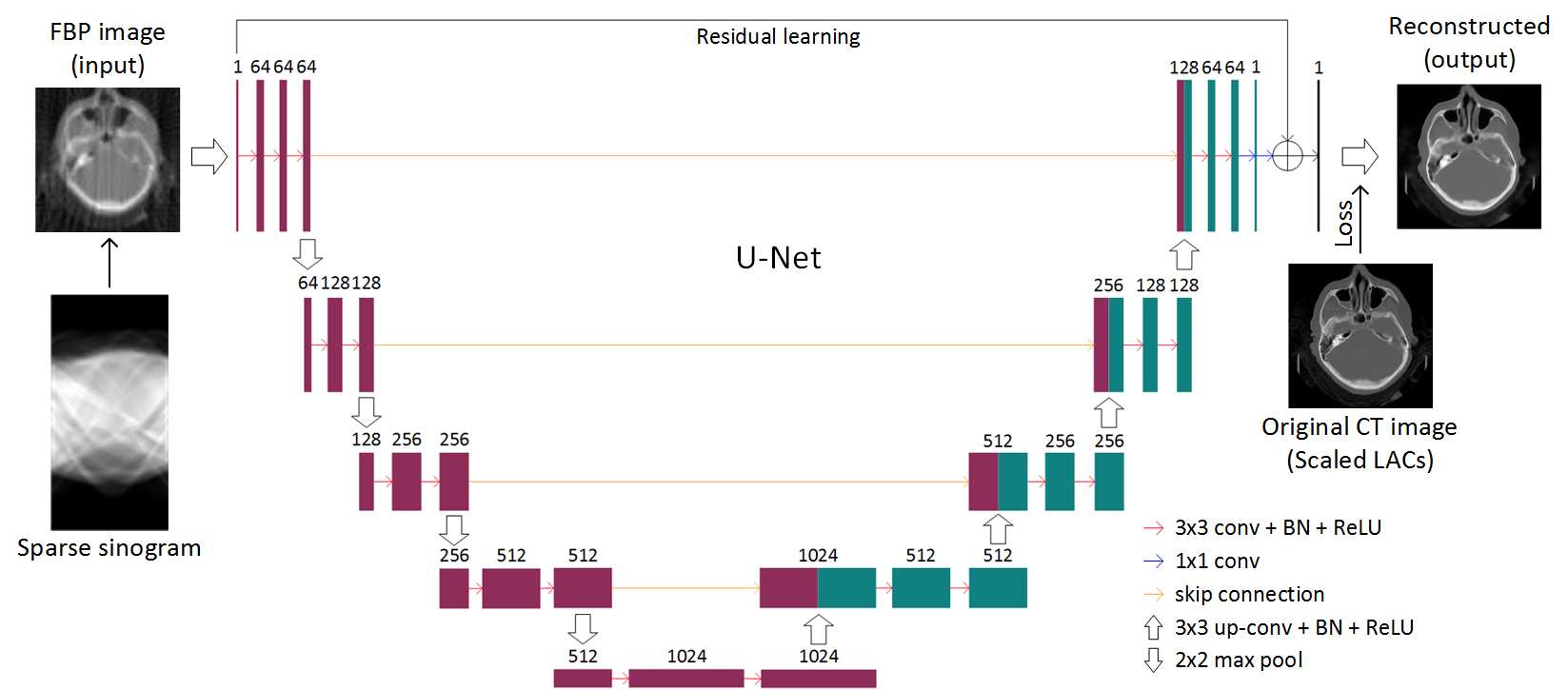}
        \caption{Network architecture of FBPConvNet for sparse image reconstruction. The FBPConvNet is a modified U-net which employs multilevel decomposition and multichannel filtering with a skip connection between input and output for residual learning. FBP, filtered backprojection; LAC, linear attenuation coeffcients}
        \label{fig:fbpconvnet}
\end{center}
\end{figure}

Reconstructed images were generated from the synthetic sinograms for models I1-I6. Original CT images were used as the reconstructed images for \textit{recon360x729} as fully sampled sinogram data could be completely reconstructed into images using filtered back projection (FBP). However, other complex algorithms are needed to reconstruct high-quality images from sparser datasets, such as model-based iterative reconstruction. Rather than employing complex iterative algorithms, we implemented a deep learning approach to reconstruct sparsely sampled sinograms as this technique has been demonstrated to compare favorably to state-of-the-art iterative algorithms for sparse-view image reconstruction \cite{jin2017deep, xie2018artifact}. We implemented FBPConvNet, a modified U-net \cite{ronneberger2015u} with multiresolution decomposition and residual learning as proposed by a prior work \cite{jin2017deep}. FBPConvNet takes FBP reconstructed images from sparser sinograms (\textit{sino120x240} or \textit{sino40x80}) as inputs and is trained for regression between the input and the original CT image (converted into LACs) with mean square error (MSE) as the loss function (Figure \ref{fig:fbpconvnet}). Since the output images of FBPConvNet were LACs, they were converted into HU as the final reconstructed images. Sparser sinograms were resized to 360x729 pixels using bilinear interpolation in order to make the corresponding FBP images have the uniform resolution of 512x512 pixels, resulting in final reconstructed images of 512x512 pixels. The best FBPConvNet models selected based on RMSE values on the validation data were employed on \textit{sino120x240} and \textit{sino40x80} to generate \textit{recon120x240} and \textit{recon40x80} respectively. The root mean square error (RMSE) of reconstructed images obtained from the FBPConvNet in validation dataset are much smaller than that of conventional FBP images (Table \ref{table:reconresults}).

\begin{figure}[!htbp]
\begin{center}
        \includegraphics[width=\linewidth]{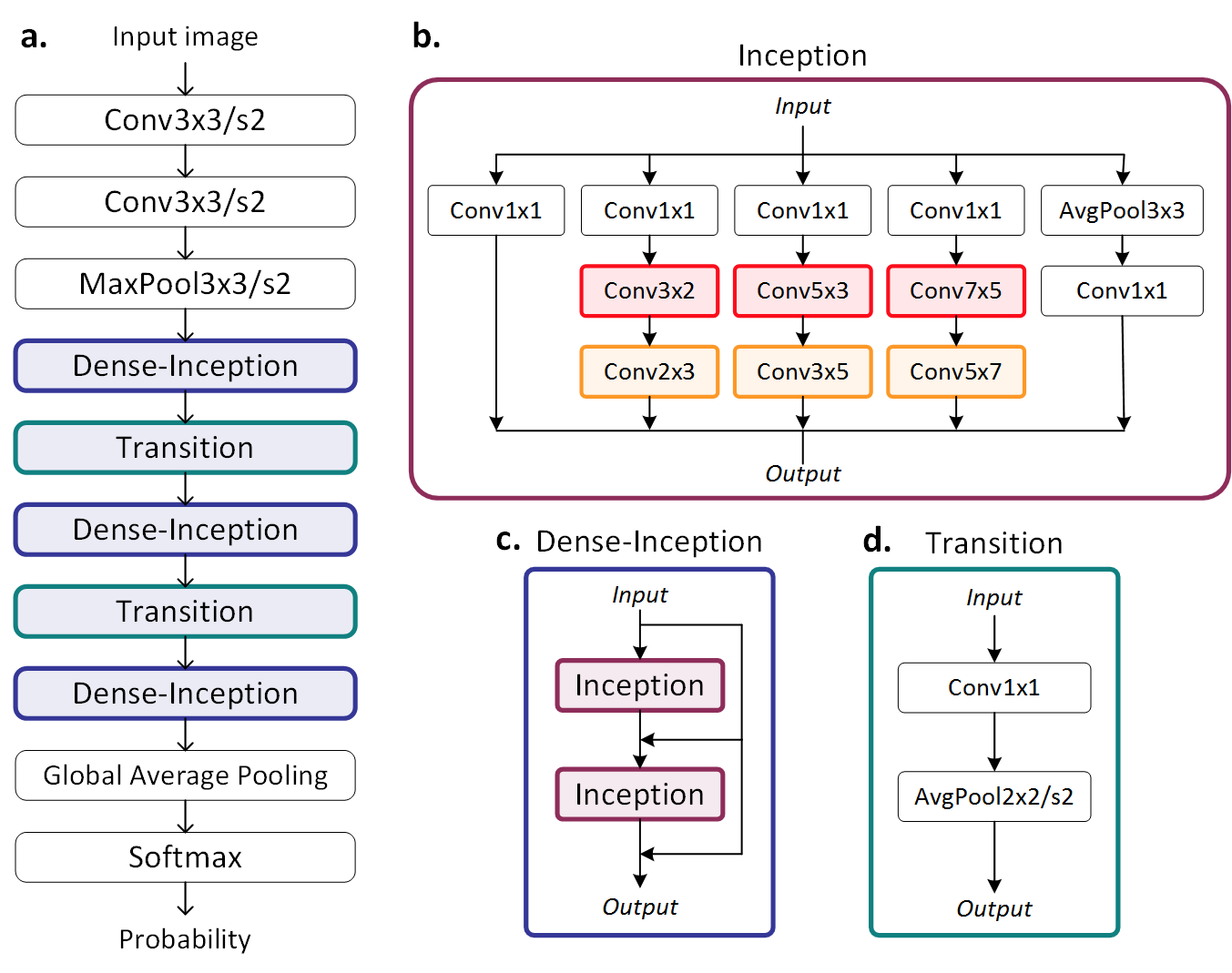}
        \caption{\textbf{a} Overall network architecture of SinoNet. \textbf{b} Detailed network diagram within the Inception modules that include rectangular convolutional filters and pooling layers. The modified Inception module contains multiple rectangular convolution filters of varying sizes: height-wise rectangular filters (projection dominant) in red; width-wise rectangular filters (detector dominant) in orange; “Conv3x3/s2” indicates a convolutional layer with 3x3 filters and 2 stride, and “Conv3x2” means a convolution layer with 3x2 filters and 1 stride. \textbf{c} Dense-Inception layers contain two densely connected Inception modules. \textbf{d} Transition modules situated between Dense-Inception modules reduce the size of feature maps. Conv = convolution layer, MaxPool = max pooling layer, AvgPool = average pooling layer}
        \label{fig:sinonet}
\end{center}
\end{figure}

\subsection{Windowed images and sinograms}

\begin{figure}[ht]
\begin{center}
        \includegraphics[width=\linewidth]{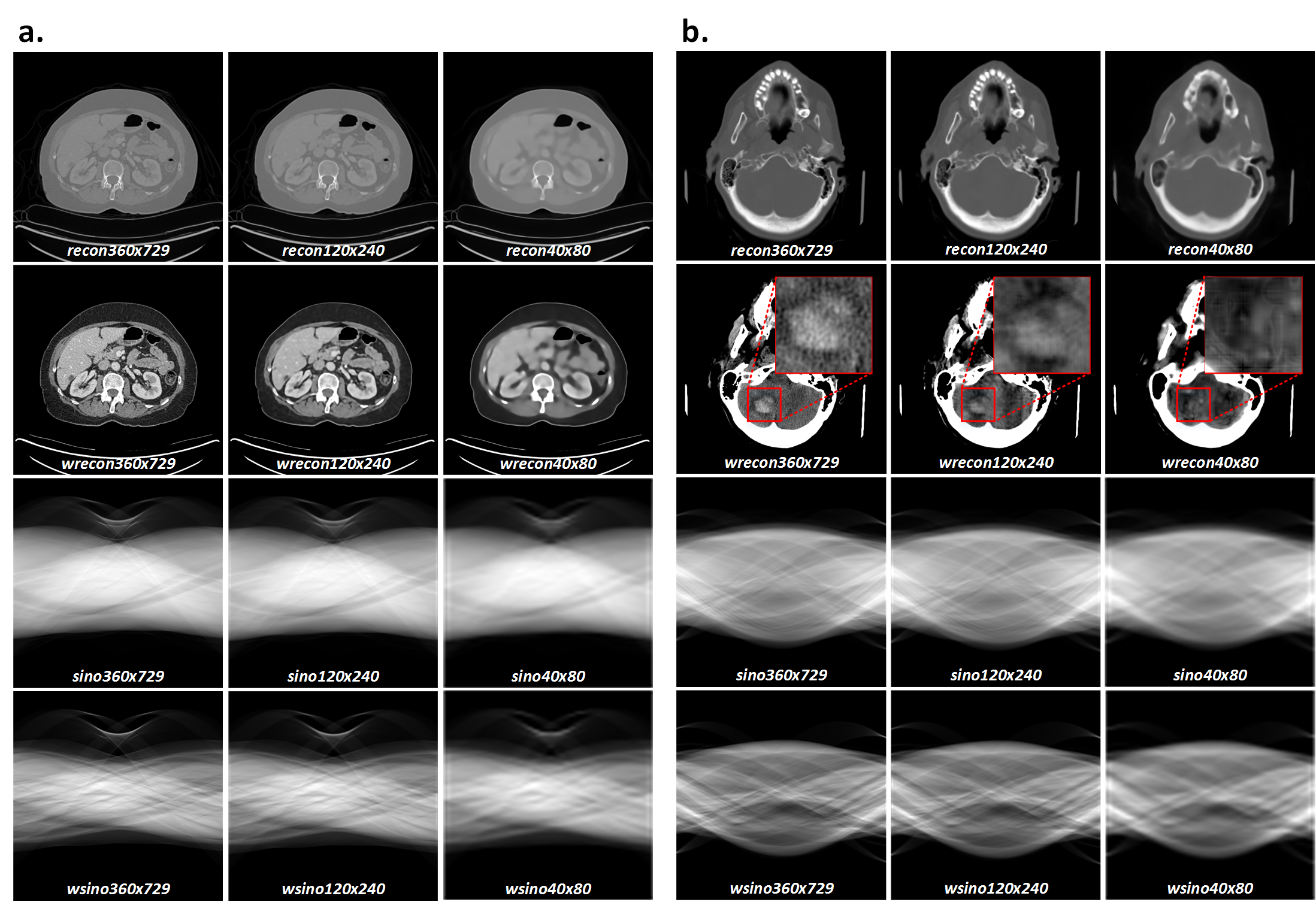}
        \caption{Example images for sinograms (\textit{sino360x729}, \textit{sino120x240}, \textit{sino40x80}) and reconstructed images (\textit{recon360x729}, \textit{recon120x240}, \textit{recon40x80}) for \textbf{a} body part recognition and \textbf{b} ICH detection tasks. Windowed reconstruction images were generated by applying abdomen window (window-level = 40 HU, window-width = 400 HU) for body part recognition and brain window (window-level = 50 HU, window-width = 100 HU) for ICH detection. All reconstructed images and sinograms are normalized to the same resolution for this figure.}
        \label{fig:examplemethods}
\end{center}
\end{figure}

We utilized full-range 12-bit grayscale images and windowed 8-bit grayscale images with different window-levels (WL) and window-widths (WW) suitable for each task: abdominal window (WL=40HU, WW=400HU) for body part recognition and brain window (WL=50HU, WW=100HU) for ICH detection. The windowed sinograms were generated from corresponding windowed CT images. Examples of windowed images and sinograms are shown in Figure \ref{fig:examplemethods}.

\subsection{Convolutional neural network for sinograms: SinoNet}

A customized convolutional neural network, SinoNet, was designed for analyzing sinograms using customized Inception modules with multiple convolutional and pooling layers and dense connection for efficient use of model parameters \cite{szegedy2016rethinking, huang2017densely}. As shown in Figure \ref{fig:sinonet}, the Inception module was modified with various sized rectangular convolutional filters in SinoNet. The non-square filters include height-wise (detector dominant) and width-wise (projection dominant) filters to enable efficient extraction of features from sinusoidal curves. Two Inception modules were densely connected to form a Dense-Inception block, which was followed by a Transition block to reduce the number and dimension of feature maps for computational efficiency, as suggested in the original report \cite{huang2017densely}. In this study, SinoNet was used only for interpreting sinograms.

\subsection{Baseline convolutional neural network: Inception-v3}

Inception-v3 \cite{szegedy2016rethinking}, a validated CNN for object recognition in the ImageNet Large Scale Visual Recognition Challenge (ILSVRC) \cite{russakovsky2015imagenet}, was selected as the network architecture to develop classification models trained on reconstructed images. We modified Inception-v3 by replacing the last fully-connected layers with a sequence of a global average pooling (GAP) layer, a fully-connected layer, and a softmax layer with outputs of the same number of categories: 16 multi-class outputs for body part recognition and a binary output for ICH detection. Inception-v3 was also used to classify sinograms when evaluating SinoNet performance at body part recognition and ICH detection when using sinograms as the input data.

\subsection{Weight initialization}

All models developed using Inception-v3 and SinoNet for body part recognition task were initialized with He normal initialization \cite{he2015delving}. For the ICH detection task, models were initialized with corresponding pre-trained weights on the body part recognition with full-view scanning geometry. For example, the Inception-v3 model trained with \textit{recon360x729} for body part recognition was used as the initial weights for Inception-v3 models trained with reconstructed images for ICH detection for all scanning geometries and window levels. Similarly, SinoNet ICH detection models were initialized using the weights from the body part recognition SinoNet model trained with \textit{sino360x729}.

\subsection{Performance evaluation and statistical analysis}

Test accuracy was used as the performance metric for comparing body part recognition models, and ROC curves with AUC were used for evaluating performance of models for detection of ICH. All performance metrics were calculated using \textit{scikit-learn 0.19.2} available in \textit{python 2.7.12}. A non-parametric approach (DeLong \cite{delong1988comparing}) was used to assess the statistical significance of the difference between AUCs of ICH detection models trained with reconstruction images and sinograms using \textit{Stata version 15.1} (StataCorp, College Station, Texas, USA). We employed a non-parametric, bootstrap approach with 2,000 iterations to compute 95\% CIs of the metrics including test accuracy and AUC \cite{efron1994introduction}.

\subsection{Network training}

Classification models for body part recognition and ICH detection were trained for 45 epochs using the Adam optimizer with default settings \cite{kingma2014adam} and a mini-batch size of 80. FBPConvNet models were trained for 100 epochs using the Adam optimizer with default settings and a mini-batch size of 20. The base learning rate of 0.001 was decayed by a factor of 10 every 15 epochs for the classification models and every 33 epochs for FBPConvNet. The best classification and FBPConvNet models were selected based on the validation loss.

\subsection{Infrastructure}

We used \emph{radon} and \emph{iradon} functions in Matlab 2018a for generating sinograms and obtaining FBP reconstructed images, respectively. We used Keras (version 2.1.1) with a Tensorflow backend (version 1.3.0) as the framework for developing deep learning models, and performed experiments using an NVIDIA Devbox (Santa Clara, CA) equipped with four TITAN X GPUs with 12GB of memory per GPU.

\newpage
\section*{Supplementary Information}

\beginsupplement

\begin{table}[!ht]
\centering
\scalebox{1.1}{
\begin{tabular}{c|c|c|c|c}
    \toprule
    % \hline
     & \multicolumn{2}{c|}{Body part recognition} & 
    \multicolumn{2}{c}{ICH detection} \\
    
    Input & FBP & FBPConvNet & 
    FBP & FBPConvNet \\
    % \hline
    \midrule
    
    \textit{sino120x240} &
    1155.4 $\pm$ 19.3 &
    28.6 $\pm$ 6.2 &
    1251.5 $\pm$ 35.6 &
    26.8 $\pm$ 8.5 \\
    
    \textit{sino40x80} &
    1147.2 $\pm$ 19.4 &
    66.9 $\pm$ 16.6 &
    1238.1 $\pm$ 34.2 &
    66.1 $\pm$ 21.2 \\
    
    \bottomrule
\end{tabular}}
\caption{RMSE computed on the validation dataset between scaled LACs converted from original CT images and reconstructed images (\textit{recon120x240}, \textit{recon40x80}) through FBP and FBPConvNet from sparse sinograms. RMSE values are expressed as mean $\pm$ standard deviation.}
\label{table:reconresults}
\end{table}

\clearpage

\bibliographystyle{unsrtnat}
\bibliography{main}

\end{document}